\documentclass[11pt]{article}

\usepackage[margin=1in]{geometry}
\usepackage{graphicx}
\usepackage{booktabs}
\usepackage{microtype}
\usepackage{amsmath,amssymb}
\usepackage{hyperref}
\usepackage{caption}
\usepackage{subcaption}

\title{UAV-Based Infrastructure Inspections: A Literature Review and Proposed Framework for AEC+FM}

\author{
Amir Farzin Nikkhah$^{1}$,
Dong Chen$^{2}$,
Bradford Campbell$^{3}$,
Somayeh Asadi$^{4}$,
Arsalan Heydarian$^{4}$
}

\date{}

\begin{document}
\maketitle

\begingroup
\renewcommand\thefootnote{\arabic{footnote}}
\footnotetext[1]{Systems and Information Engineering, University of Virginia, Charlottesville, VA 22903, USA.
Email: \texttt{kgc2mj@virginia.edu}}

\footnotetext[2]{Agricultural \& Biological Engineering, Mississippi State University, Starkville, MS 39762, USA.
Email: \texttt{dc2528@misstate.edu}}

\footnotetext[3]{Link Lab, University of Virginia, Charlottesville, VA 22903, USA.
Email: \texttt{bradjc@virginia.edu}}

\footnotetext[4]{Link Lab \& Civil and Environmental Engineering, University of Virginia, Charlottesville, VA 22903, USA.
Emails: \texttt{kn3gr@virginia.edu, ah6rx@virginia.edu}}
\endgroup

\begin{center}
\small
\textbf{Note:} This is a preprint of a paper accepted for publication at the
\emph{International Conference on Construction Engineering and Management (I3CE 2025)}.

\end{center}

\begin{abstract}
Unmanned Aerial Vehicles (UAVs) are transforming infrastructure inspections in the Architecture, Engineering, Construction, and Facility Management (AEC+FM) domain.
By synthesizing insights from over 150 studies, this review paper highlights UAV-based methodologies for data acquisition, photogrammetric modeling, defect detection, and decision-making support.
Key innovations include path optimization, thermal integration, and advanced machine learning (ML) models such as YOLO and Faster R-CNN for anomaly detection.
UAVs have demonstrated value in structural health monitoring (SHM), disaster response, urban infrastructure management, energy efficiency evaluations, and cultural heritage preservation.
Despite these advancements, challenges in real-time processing, multimodal data fusion, and generalizability remain.
A proposed workflow framework, informed by literature and a case study, integrates RGB imagery, LiDAR, and thermal sensing with transformer-based architectures to improve accuracy and reliability in detecting structural defects, thermal anomalies, and geometric inconsistencies.
The proposed framework ensures precise and actionable insights by fusing multimodal data and dynamically adapting path planning for complex environments, presented as a comprehensive step-by-step guide to address these challenges effectively.
This paper concludes with future research directions emphasizing lightweight AI models, adaptive flight planning, synthetic datasets, and richer modality fusion to streamline modern infrastructure inspections.
\end{abstract}

\section{Introduction}

Unmanned Aerial Vehicles (UAVs) have increasingly supported a variety of industries by providing safe, cost-effective, and efficient means of collecting and analyzing data.
In the AEC+FM sector, traditional inspection methods often involve visual inspections performed by teams of experts, which can be labor-intensive, time-consuming, and costly \cite{shin2023}.
Moreover, these manual inspections pose safety risks \cite{ko2021}.
UAVs offer a safer and more efficient alternative by enabling remote inspections \cite{yu2023}.
They can survey large areas rapidly, provide detailed imagery, and integrate computational tools to detect defects and anomalies.

Studies have shown that UAV inspections can be completed in a fraction of the time required by traditional methods, often within 10--20\% of duration \cite{wang2024b}, and can reduce inspection costs by up to 40--60\% for infrastructure projects \cite{buffi2017}.
Concurrently, UAV integration with advanced analytics---such as convolutional neural networks (CNNs) for feature extraction \cite{saeed2021,yoon2024}, transformer-based architectures for occluded object detection \cite{ye2023}, and real-time data processing---enables automated and precise detection of structural issues.
Integration with Building Information Modeling (BIM) and Geographic Information Systems (GIS) has further advanced the field, enabling digital twins and interactive three-dimensional models \cite{shin2023}.

UAV-based inspections support diverse applications, including structural health monitoring (SHM) \cite{yoon2024}, disaster response \cite{levine2022}, and energy efficiency evaluation \cite{pan2020}. Furthermore, Jofré‐briceño et al. \cite{jofrebriceno2021} combined UAV photogrammetry with BIM to automate port infrastructure maintenance workflows.
Despite progress, challenges persist, including limited flight time, weather sensitivity, privacy considerations, and data alignment complexities.
Defect detection models, while increasingly accurate, can be computationally demanding for real-time applications and may lack generalizability across diverse infrastructure types and environmental conditions.
While prior studies have examined UAV-based inspection approaches \cite{ko2021,shin2023}, this work focuses on multi-sensor fusion and emerging deep learning frameworks, reviewing relevant implementations and proposing an integrated inspection framework.

This review examines how existing research leverages UAVs for inspecting various infrastructure types, emphasizing gaps such as path planning optimization and machine learning-based multi-sensor fusion.
To address these gaps, the paper proposes a comprehensive framework for multi-sensor UAV-based infrastructure inspections.

\section{Literature Review}

A semi-systematic literature review was conducted by querying the Scopus database using terms such as ``UAV photogrammetry,'' ``drone inspections,'' ``AEC+FM,'' and ``machine learning for infrastructure inspection,'' covering publications from 2017 to 2024.
The initial search yielded approximately 300 articles.
Two inclusion criteria were applied: (i) studies explicitly addressing UAV-based data collection or defect detection, and (ii) relevance to AEC+FM applications.
Screening abstracts reduced the pool to 150 articles, which were subsequently reviewed in full.
The analysis focused on flight path planning, sensor integration, data processing pipelines, and reported outcomes or case studies.

Photogrammetric configurations such as 66.7\% forward overlap and 50\% side overlap have been shown to improve three-dimensional reconstruction accuracy \cite{wang2022}.
Optimized UAV flight planning can significantly reduce processing time for bridge inspections \cite{wang2024b}.
Convex Hull and Bat Algorithm approaches streamline large-scale inspections by reducing mission duration \cite{biundini2020}.
Voronoi-based planning improves obstacle avoidance and safety in complex environments \cite{adachi2023}, while Ant Colony Optimization enhances coverage efficiency \cite{yu2023}.
Quality-driven path refinement further improves reconstruction completeness \cite{zhang2024}.

Deep learning models have enabled automated defect detection.
YOLO-based pipelines achieve high crack detection accuracy \cite{woo2023}.
Faster R-CNN models excel in facade damage detection \cite{shin2023}, while cascade detection methods improve small-defect identification \cite{ying2022}.
Real-time frameworks such as RTD-Net combine CNNs and transformers to achieve robust performance 86\% accuracy \cite{ye2023}, and in-flight image quality assessment methods reduce post-processing overhead \cite{wang2024a}.

UAV photogrammetry combined with Structure-from-Motion (SfM) techniques achieves millimeter-level crack measurement accuracy \cite{ioli2022}.
Open-source frameworks integrated with consumer-grade UAVs provide cost-effective inspection tools, achieving over 80\% crack detection accuracy \cite{ko2021}.
However, effective integration of LiDAR and thermal sensors with automated detection models remains underexplored \cite{wang2022}, highlighting the need for improved multimodal fusion strategies. Further recent studies highlight that additional research is needed to develop faster and more efficient methods for processing and analyzing multi-sensor data, thereby enhancing the accuracy and scalability of real-time inspections \cite{wang2024a}.

\section{Proposed Framework}

\begin{figure}[t]
    \centering
    \includegraphics[width=0.95\linewidth]{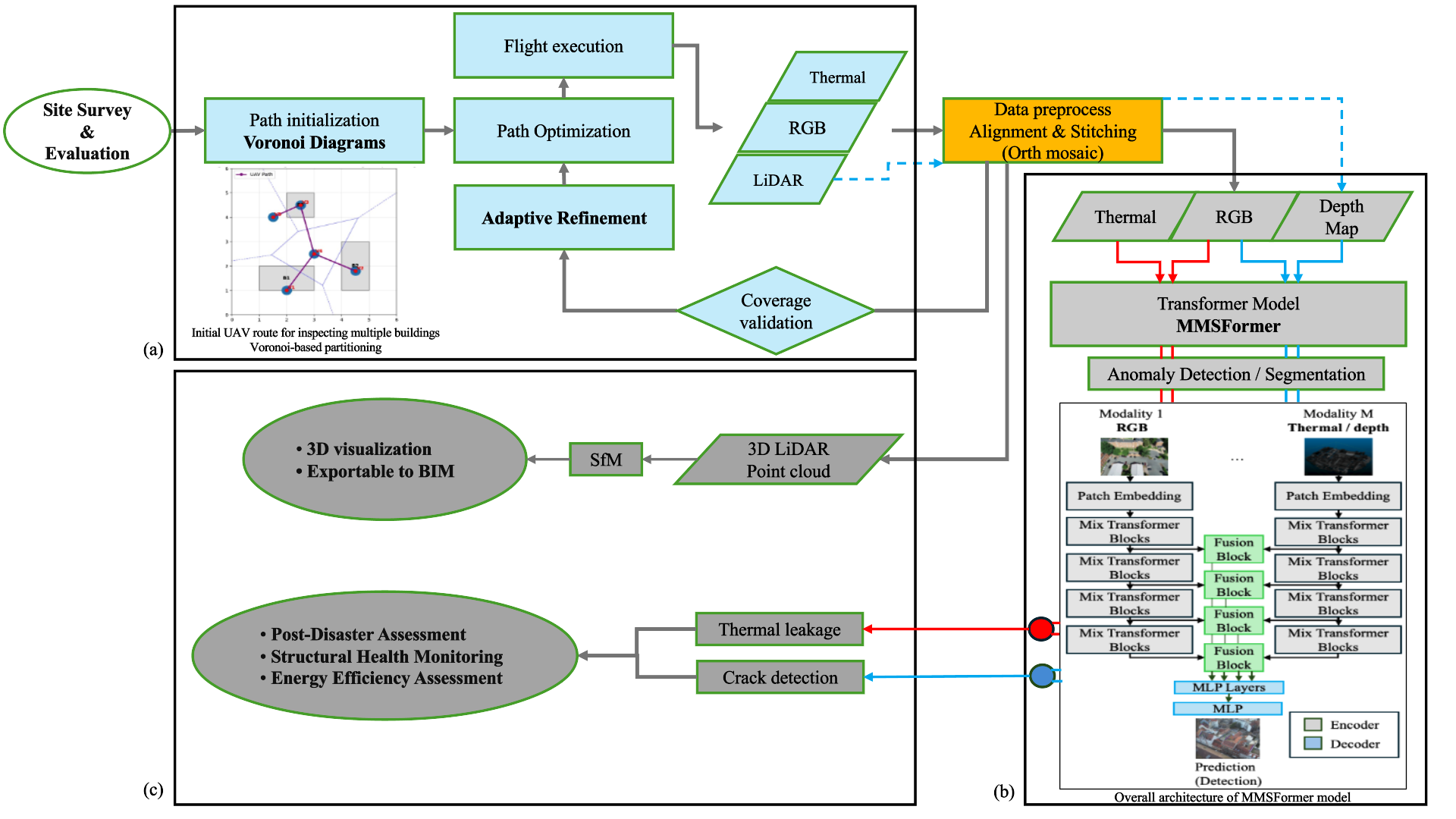}
    \caption{Proposed UAV-Based Inspection Framework: (a) Path optimization, (b) data processing with MMSFormer Transformer model, and (c) output use-cases and applications.}
    \label{fig:framework}
\end{figure}

Building on the insights from the literature review, this section presents a UAV-based
inspection framework tailored for AEC+FM industry. The framework integrates rigorous path
planning algorithms to cover maximum area with dynamic path correction, multi-sensor data
acquisition (RGB, LiDAR, thermal), advanced multimodal data fusion, and state-of-the-art defect
detection architecture framework for aerial taken images. A case study at the University of
Virginia’s main campus in Charlottesville, VA, exemplifies the approach, demonstrating how
careful mission planning, robust data processing, and cutting-edge deep learning architectures can
streamline infrastructure assessment tasks. Figure~\ref{fig:framework} outlines the step-by-step framework, from defining objectives to actionable outcomes.

\subsection{Study Area and Data Acquisition Setup}

The selected test area encompasses multiple academic buildings and facilities on the UVA grounds. These structures vary in height, facade complexity, and material composition, making them ideal for testing multi-sensor UAV inspections. Initial data collection campaigns were conducted with a DJI Matrice 300 RTK UAV equipped with a LiDAR sensor and an RGB camera at altitudes of approximately 60 m and 80 m. Flights occurred over a span of four days, during morning (7–9 AM) and late afternoon (5–7 PM) windows, capturing diverse lighting conditions and environmental factors such as changing temperatures, and varying cloud cover. 

Weather conditions during the experiments included average daytime temperatures of approximately 90$^\circ$F and varying wind speeds between 2.2 to 13.4 mph, with gusts up to 35.7 mph. These dynamic conditions underscore the importance of both robust flight path planning and adaptive inspection strategies. The chosen forward overlap of ~60\% and side overlap of ~50\% 
ensured sufficient redundancy for photogrammetric processing, while LiDAR scans improved geometric accuracy. Future data acquisitions will incorporate thermal cameras to facilitate thermal anomaly detection (e.g., insulation issues and heat leaks) and enrich the multimodal dataset.

\subsection{Path Design and Optimization Method}

Comprehensive and efficient data coverage is critical for accurate defect detection and quality 3D reconstructions. Conventional ``lawnmower'' patterns suffice for uniform areas, but complex urban environments with irregular building layouts require more adaptive methods. We propose a quality-driven, Voronoi-based path optimization approach, further refined by incorporating building geometry and priority zones identified through BIM or GIS data 
layers \cite{adachi2023}.

For the initial path generation for inspection, the area is delineated, and key structures are identified. Voronoi diagrams are generated based on building footprints and known obstacles, providing a partitioned representation of the space. By referencing these polygons, the UAV’s initial waypoints ensure coverage of critical surfaces (e.g., facades, rooftops) while minimizing redundant passes. A quality-driven method \cite{zhang2024} evaluates initial image sets for coverage gaps, poor lighting conditions, or missed facade elements for an adaptive refinement. 
Adjustments to flight altitude and camera tilt can be made mid-mission, guided by real-time feedback systems integrated into the UAV’s controller interface. By integration with BIM/CAD Data, where it is available, overlaying these data on top of the Voronoi layout, the path is fine
tuned to target structural details essential for defect detection, ensuring that areas prone to cracking, spalling, or thermal anomalies receive more focused attention. This combined Voronoi and quality-driven approach ensures robust coverage, reduces unnecessary flights, and improves data 
quality, ultimately supporting more accurate and reliable defect assessments. 

Upon completing data acquisition, RGB images, LiDAR point clouds, and thermal imagery (in future acquisitions in our case) undergo preprocessing. Structure-from-Motion (SfM) and Multi-View Stereo (MVS) techniques reconstruct detailed 3D models from RGB images, while LiDAR point clouds provide precise geometric references and height maps. Thermal data is aligned with the RGB-derived orthomosaics and integrated into the 3D model, producing a “thermal point cloud” that combines surface details with subsurface anomalies. The fused datasets yield a rich multimodal representation, enabling more over toned and accurate defect detection. Thermal imaging helps identify insulation failure, moisture intrusion, or heat leaks, while LiDAR data ensures the spatial fidelity of the reconstructed environment.

\subsection{Defect Detection and Thermal Leak Assessment}

Defect detection is performed using a multimodal transformer-based framework inspired by MMSFormer \cite{reza2024}.
It is demonstrated across all three datasets; it achieves higher 
segmentation accuracy and better robustness in multimodal scenarios than common baseline methods. MMSFormer leverages modality-specific encoders and hierarchical fusion blocks to integrate data from multiple sources—originally RGB, thermal, and depth. In our adaptation, we use RGB+Thermal to enhance the detection of thermal anomalies and RGB+Height (from LiDAR) to improve spatial accuracy in identifying structural defects such as cracks or delamination. 

Modality-Specific Encoders process separate each input modality, RGB, thermal, and height maps. Hierarchical Fusion Blocks intermediate features from each encoder pass through a transformer-based fusion mechanism, aligning semantic and geometric information. The final output layers produce segmentation maps and defect classifications, allowing for localized detection of cracks, corrosion, or thermal inefficiencies. This method is expected to outperform traditional single-modality methods by employing MMSFormer or a similar state-of-the-art transformer-based segmentation model, achieving higher intersection-over-union (IoU) and mean accuracy metrics. This robust approach is well-suited for complex environments like campus buildings, where structures vary widely in geometry, material, and environmental conditions. 

\section{Discussion}

\begin{figure}[t]
    \centering
    \includegraphics[width=0.85\linewidth]{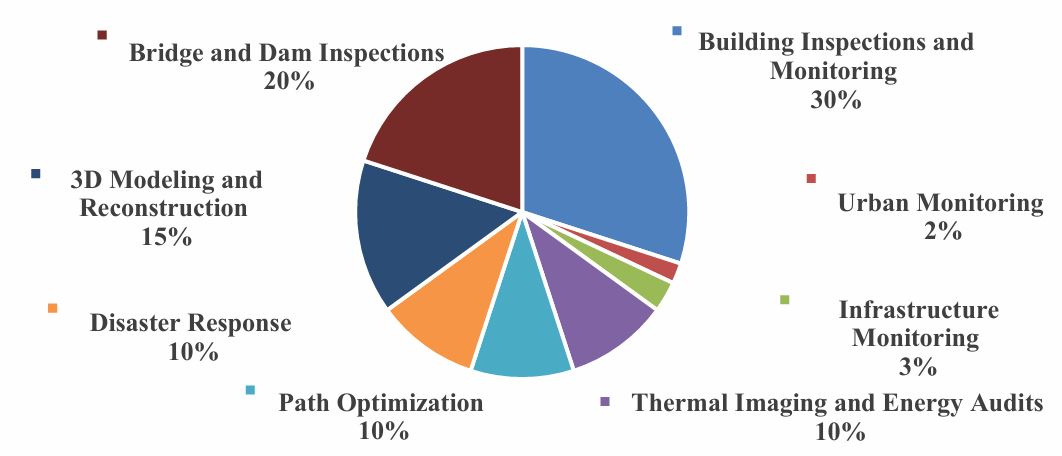}
    \caption{Distribution of UAV-based applications in infrastructure inspections.}
    \label{fig:applications}
\end{figure}

Based on reviewed studies, various applications in AEC+FM have different methods that best suit the goal in terms of accuracy, time, and complexity. For instance, SHM-oriented flights may emphasize high-resolution imaging and precise defect-detection models to locate microcracks or delamination, even if this requires longer flight times and more substantial processing. In contrast, energy-focused inspections might prioritize integrating thermal cameras and covering large areas rapidly to identify heat loss, thereby reducing flight duration but accepting slightly lower imaging detail. Hence, the first step in aerial inspections for infrastructure should be the definition of the goal. Figure~\ref{fig:applications}  presents the percentage of studies focused on each application area, illustrating UAVs’ broad impact across SHM, disaster response, energy/urban monitoring, cultural heritage, and facility management. Furthermore, our reviews highlighted limited studies have evaluated real world constraints such as various weather conditions, time of day and natural lighting availability, as well as integration of multi-sensor data such as LiDAR, RGB and thermal for real-time decision making. These gaps highlight the need for a more holistic, adaptable framework, one that accommodates diverse environmental settings and sensor modalities while preserving high levels of efficiency and accuracy. 

Advanced optimization methods like Voronoi diagrams, Ant Colony Optimization (ACO), and Convex Hull-based approaches improve UAV inspection efficiency by ensuring comprehensive coverage while minimizing redundancy. ACO and lawnmower patterns streamline systematic area inspections, while Voronoi-based planning dynamically adapts to complex environments, prioritizing critical zones. Integrating BIM/CAD data into path planning further refines flight routes, ensuring coverage of high-priority structural areas. Quality-driven strategies, such as tilt photogrammetry and optimized overlap settings, enhance 3D reconstruction fidelity, reducing errors and improving model completeness. While these techniques advance inspection workflows, their computational requirements can limit real-time adaptability. 

The reviewed studies highlight that the choice of the CNN model depends on inspection goals. As summarized in Table~\ref{tab:model_comparison}, for high-speed, large-area inspections, YOLO is optimal due to its single-stage detection framework, while two-stage models like Faster R-CNN and Mask RCNN excel in accuracy-critical tasks detecting smaller or complex defects, such as microcracks or 
delaminations. SSD offers a multi-scale feature map that enables efficient detection of objects of varying sizes, providing flexibility in defect detection. It is ideal for routine UAV inspections, residential building inspections, and urban damage mapping. Transformer-based architectures like 
RTD-Net and MMSFormer combine speed and precision, offering higher accuracy and computational efficiency for multimodal inputs. 

\begin{table}[t]
\centering
\caption{Comparison of potential models for UAV-based inspections.}
\label{tab:model_comparison}
\begin{tabular}{p{3cm} p{3cm} p{3cm} p{5cm}}
\toprule
Model & Speed & Accuracy & Applications / Best use case \\
\midrule
YOLO & High (real-time) & Moderate & Large-scale inspections, disaster response, safety patrols \\
Faster / Mask R-CNN & Low (non-real-time) & High & Detailed facade segmentation, post-event assessment, SHM \\
SSD & Moderate & Moderate--High & Routine maintenance and urban inspections \\
MMSFormer & Moderate--High & Very high & Multimodal fusion in complex environments \\
\bottomrule
\end{tabular}
\end{table}

The proposed workflow tackles key challenges in drone-based inspections, including the effective integration of multi-sensor processing, optimized coverage in complex environments, inspecting areas with diverse facades, and the efficient utilization of LiDAR and thermal sensors in aerial applications. This is achieved through a seamless inspection pipeline that combines path optimization, multi-sensor data fusion, and AI-driven defect detection.

Combining Voronoi-based algorithms with semi-real-time feedback ensures robust flight coverage, even in complex urban settings. Multi-modal fusion of RGB, height maps, and thermal imaging provides a comprehensive 3D representation, bridging surface-level and subsurface defect analysis. Transformer-based MMSFormer models process multimodal inputs (RGB+Thermal or 
RGB+Height), improving segmentation accuracy for cracks, delaminations, and thermal anomalies. Detected defects and thermal anomalies are overlaid on interactive 3D models, facilitating decision-making for maintenance and risk assessment. 

\section{Conclusion and Future Work}

Key challenges include dataset limitations, labor-intensive labeling, and environmental factors like lighting variability and occlusions. Computationally heavy models, such as Mask RCNN, hinder real-time operations due to latency and power consumption. Despite advancements, limitations remain in dataset diversity, environmental variability (lighting and occlusions), and computational efficiency of defect detection models. Solutions include developing synthetic datasets for training with complex CNN or transformer-based models, implementing lightweight models for real-time UAV processing, enhancing adaptive flight planning through real-time 
feedback systems, and advancing multimodal fusion techniques (RGB + Thermal + LiDAR) for complex environments. 
Future efforts should focus on lightweight AI models, synthetic datasets, adaptive flight planning, integration of LiDAR sensors, and the development of faster multimodal fusion models for real-time applications to address the gaps. Moreover, integrating UAV-generated data with interior-building sensor information, such as HVAC system outputs or indoor environmental 
measurements, can offer a more holistic view of structural and thermal performance, enabling stakeholders to make more informed decisions about maintenance, retrofitting, and occupant comfort. Future efforts can focus on refining these multi-sensors inside-out approaches to enhance accuracy, reduce operational costs, and expand real-time applicability across diverse infrastructure types and scenarios.

\bibliographystyle{plain}
\bibliography{refs}

@inproceedings{adachi2023,
  author    = {Adachi, K. and Miwa, H.},
  title     = {Inspection Method of Building Surface Condition by Unmanned Aerial Vehicle Under Challenged Communication Environment},
  booktitle = {International Conference on Intelligent Networking and Collaborative Systems},
  pages     = {1--13},
  publisher = {Springer},
  year      = {2023}
}

@incollection{biundini2020,
  author    = {Biundini, I. Z. and Melo, A. G. and Pinto, M. F. and Marins, G. M. and Marcato, A. L. M. and Honorio, L. M.},
  title     = {Coverage Path Planning Optimization for Slopes and Dams Inspection},
  booktitle = {Advances in Intelligent Systems and Computing},
  pages     = {513--523},
  publisher = {Springer},
  year      = {2020}
}

@article{buffi2017,
  author  = {Buffi, G. and Manciola, P. and Grassi, S. and Barberini, M. and Gambi, A.},
  title   = {Survey of the Ridracoli Dam: UAV--based photogrammetry and traditional topographic techniques in the inspection of vertical structures},
  journal = {Geomatics, Natural Hazards and Risk},
  year    = {2017},
  publisher = {Taylor \& Francis}
}

@inproceedings{ioli2022,
  author    = {Ioli, F. and Pinto, A. and Pinto, L.},
  title     = {UAV Photogrammetry for Metric Evaluation of Concrete Bridge Cracks},
  booktitle = {International Archives of the Photogrammetry, Remote Sensing and Spatial Information Sciences (ISPRS Archives)},
  pages     = {1025--1032},
  publisher = {International Society for Photogrammetry and Remote Sensing},
  year      = {2022}
}

@article{jofrebriceno2021,
  author  = {Jofr{\'e}-brice{\~n}o, C. and La Rivera, F. M. and Atencio, E. and Herrera, R. F.},
  title   = {Implementation of facility management for port infrastructure through the use of uavs, photogrammetry and bim},
  journal = {Sensors},
  volume  = {21},
  number  = {19},
  year    = {2021},
  publisher = {MDPI},
  doi     = {10.3390/s21196686}
}

@inproceedings{ko2021,
  author    = {Ko, P. and Prieto, S. A. and de Soto, B. G.},
  title     = {ABECIS: an Automated Building Exterior Crack Inspection System using UAVs, Open-Source Deep Learning and Photogrammetry},
  booktitle = {Proceedings of the International Symposium on Automation and Robotics in Construction},
  pages     = {637--644},
  publisher = {International Association for Automation and Robotics in Construction (IAARC)},
  year      = {2021}
}

@article{levine2022,
  author  = {Levine, N. M. and Narazaki, Y. and Spencer, B. F.},
  title   = {Performance-based post-earthquake building evaluations using computer vision-derived damage observations},
  journal = {Advances in Structural Engineering},
  volume  = {25},
  number  = {16},
  pages   = {3425--3449},
  year    = {2022},
  publisher = {SAGE Publications},
  doi     = {10.1177/13694332221119883}
}

@article{pan2020,
  author  = {Pan, N. H. and Tsai, C. H. and Chen, K. Y. and Sung, S.},
  title   = {Improvement of uav based an evaluation approach to mid-high rise buildings’ exterior walls},
  journal = {SDHM Structural Durability and Health Monitoring},
  volume  = {14},
  number  = {2},
  pages   = {109--125},
  year    = {2020},
  publisher = {Tech Science Press},
  doi     = {10.32604/SDHM.2020.06489}
}

@article{reza2024,
  author  = {Reza, M. K. and Prater-Bennette, A. and Asif, M. S.},
  title   = {MMSFormer: Multimodal Transformer for Material and Semantic Segmentation},
  journal = {IEEE Open Journal of Signal Processing},
  volume  = {5},
  pages   = {599--610},
  year    = {2024},
  publisher = {IEEE},
  doi     = {10.1109/OJSP.2024.3389812}
}

@inproceedings{saeed2021,
  author    = {Saeed, M. S.},
  title     = {Unmanned Aerial Vehicle for Automatic Detection of Concrete Crack using Deep Learning},
  booktitle = {International Conference on Robotics, Electrical and Signal Processing Techniques},
  pages     = {624--628},
  year      = {2021}
}

@article{shin2023,
  author  = {Shin, H. and Kim, J. and Kim, K. and Lee, S.},
  title   = {Empirical Case Study on Applying Artificial Intelligence and Unmanned Aerial Vehicles for the Efficient Visual Inspection of Residential Buildings},
  journal = {Buildings},
  volume  = {13},
  number  = {11},
  year    = {2023},
  publisher = {MDPI},
  doi     = {10.3390/buildings13112754}
}

@article{wang2024a,
  author  = {Wang, F. and Zou, Y. and Chen, X. and Zhang, C. and Hou, L. and del Rey Castillo, E. and Lim, J. B. P.},
  title   = {Rapid in-flight image quality check for UAV-enabled bridge inspection},
  journal = {ISPRS Journal of Photogrammetry and Remote Sensing},
  volume  = {212},
  pages   = {230--250},
  year    = {2024},
  publisher = {Elsevier},
  doi     = {10.1016/j.isprsjprs.2024.05.008}
}

@article{wang2024b,
  author  = {Wang, F. and Zou, Y. and del Rey Castillo, E. and Ding, Y. and Xu, Z. and Zhao, H. and Lim, J. B. P.},
  title   = {Automated UAV path-planning for high-quality photogrammetric 3D bridge reconstruction},
  journal = {Structure and Infrastructure Engineering},
  volume  = {20},
  number  = {10},
  pages   = {1595--1614},
  year    = {2024},
  publisher = {Taylor \& Francis},
  doi     = {10.1080/15732479.2022.2152840}
}

@article{wang2022,
  author  = {Wang, F. and Zou, Y. and Del Rey Castillo, E. and Lim, J. B. P.},
  title   = {Optimal UAV Image Overlap for Photogrammetric 3D Reconstruction of Bridges},
  journal = {IOP Conference Series: Earth and Environmental Science},
  year    = {2022},
  publisher = {Institute of Physics}
}

@article{woo2023,
  author  = {Woo, H. J. and Hong, W. H. and Oh, J. and Baek, S. C.},
  title   = {Defining Structural Cracks in Exterior Walls of Concrete Buildings Using an Unmanned Aerial Vehicle},
  journal = {Drones},
  volume  = {7},
  number  = {3},
  year    = {2023},
  publisher = {MDPI},
  doi     = {10.3390/drones7030149}
}

@article{ye2023,
  author  = {Ye, T. and Qin, W. and Zhao, Z. and Gao, X. and Deng, X. and Ouyang, Y.},
  title   = {Real-Time Object Detection Network in UAV-Vision Based on CNN and Transformer},
  journal = {IEEE Transactions on Instrumentation and Measurement},
  volume  = {72},
  year    = {2023},
  publisher = {IEEE},
  doi     = {10.1109/TIM.2023.3241825}
}

@inproceedings{ying2022,
  author    = {Ying, K. and Wang, J.},
  title     = {A Novel Sample Labelling Criterion for Pin Defect Detection in UAV Image},
  booktitle = {International Archives of the Photogrammetry, Remote Sensing and Spatial Information Sciences (ISPRS Archives)},
  pages     = {237--242},
  publisher = {International Society for Photogrammetry and Remote Sensing},
  year      = {2022}
}

@article{yoon2024,
  author  = {Yoon, J. and Shin, H. and Kim, K. and Lee, S.},
  title   = {CNN- and UAV-Based Automatic 3D Modeling Methods for Building Exterior Inspection},
  journal = {Buildings},
  volume  = {14},
  number  = {1},
  year    = {2024},
  publisher = {MDPI},
  doi     = {10.3390/buildings14010005}
}

@article{yu2023,
  author  = {Yu, L. and Huang, M. M. and Jiang, S. and Wang, C. and Wu, M.},
  title   = {Unmanned aircraft path planning for construction safety inspections},
  journal = {Automation in Construction},
  volume  = {154},
  year    = {2023},
  publisher = {Elsevier},
  doi     = {10.1016/j.autcon.2023.105005}
}

@article{zhang2024,
  author  = {Zhang, S. and Liu, C. and Haala, N.},
  title   = {Guided by model quality: UAV path planning for complete and precise 3D reconstruction of complex buildings},
  journal = {International Journal of Applied Earth Observation and Geoinformation},
  volume  = {127},
  year    = {2024},
  publisher = {Elsevier},
  doi     = {10.1016/j.jag.2024.103667}
}

\end{document}